\pdfoutput=1

\documentclass[letterpaper]{article} %
  \usepackage{geometry}
  \usepackage[T1]{fontenc}
  \usepackage{newtxtext}
  \usepackage{helvet}
  \usepackage{courier}
  \usepackage{xcolor}
  \AtBeginDocument{%
    \let\cite\citep%
    \bibliographystyle{abbrvnat}%
  }
  \newenvironment{links}%
    {\begin{flushleft}\setlength{\parskip}{2pt}%
     \newcommand{\link}[2]{\par\textbf{##1} --- \url{##2}}}%
    {\end{flushleft}\vskip 1ex}
  \makeatletter
  \let\@dblfloat\@float
  \let\end@dblfloat\end@float
  \makeatother
  \date{}
  \usepackage{adjustbox}

\newcommand{\cv}[1]{}
\newcommand{\av}[1]{#1}

\usepackage{comment}
\excludecomment{cvblock}\includecomment{avblock}

\usepackage[hyphens]{url}  %
\usepackage{graphicx} %
\usepackage{natbib}  %
\usepackage{caption} %
\usepackage{algorithm}
\usepackage{algorithmic}

\usepackage{newfloat}
\usepackage{listings}
\DeclareCaptionStyle{ruled}{labelfont=normalfont,labelsep=colon,strut=off} %
\floatstyle{ruled}
\newfloat{listing}{tb}{lst}{}
\floatname{listing}{Listing}

\usepackage{amssymb}            %
\definecolor{keywordcolor}{rgb}{0.7, 0.1, 0.1}   %
\definecolor{tacticcolor}{rgb}{0.0, 0.1, 0.6}    %
\definecolor{commentcolor}{rgb}{0.4, 0.4, 0.4}   %
\definecolor{symbolcolor}{rgb}{0.0, 0.1, 0.6}    %
\definecolor{sortcolor}{rgb}{0.1, 0.5, 0.1}      %
\definecolor{attributecolor}{rgb}{0.7, 0.1, 0.1} %
\lstdefinelanguage{lean} {
mathescape=false,
texcl=false,
morekeywords=[1]{
import, prelude, protected, private, noncomputable, definition, meta, renaming,
hiding, parameter, parameters, begin, constant, constants,
lemma, variable, variables, theory,
print, theorem, example,
open, as, export, override, axiom, axioms, inductive, with,
structure, record, universe, universes,
alias, help, precedence, reserve, declare_trace, add_key_equivalence,
match, infix, infixl, infixr, notation, postfix, prefix, instance,
reduce,
end, this,
using, using_well_founded, namespace, section,
attribute, local, set_option, extends, include, omit, class,
raw, replacing,
calc, have, show, suffices, by, in, at, let, forall, Pi, fun,
exists, if, dif, then, else, assume, obtain, from, register_simp_ext, unless, break, continue,
mutual, do, def, run_cmd, const,
partial, mut, where, macro, syntax, deriving,
return, try, catch, for, macro_rules, declare_syntax_cat, abbrev},
morekeywords=[2]{Sort, Type, Prop},
morekeywords=[3]{
assumption,
intro, intros, allGoals,
generalize, clear, revert, done, exact,
refine, repeat, cases, rewrite, rw,
simp, simp_all, contradiction,
constructor, injection,
induction,
},
literate=
{α}{{\ensuremath{\mathrm{\alpha}}}}1
{β}{{\ensuremath{\mathrm{\beta}}}}1
{γ}{{\ensuremath{\mathrm{\gamma}}}}1
{δ}{{\ensuremath{\mathrm{\delta}}}}1
{ε}{{\ensuremath{\mathrm{\varepsilon}}}}1
{ζ}{{\ensuremath{\mathrm{\zeta}}}}1
{η}{{\ensuremath{\mathrm{\eta}}}}1
{θ}{{\ensuremath{\mathrm{\theta}}}}1
{ι}{{\ensuremath{\mathrm{\iota}}}}1
{κ}{{\ensuremath{\mathrm{\kappa}}}}1
{μ}{{\ensuremath{\mathrm{\mu}}}}1
{ν}{{\ensuremath{\mathrm{\nu}}}}1
{ξ}{{\ensuremath{\mathrm{\xi}}}}1
{π}{{\ensuremath{\mathrm{\mathnormal{\pi}}}}}1
{ρ}{{\ensuremath{\mathrm{\rho}}}}1
{σ}{{\ensuremath{\mathrm{\sigma}}}}1
{τ}{{\ensuremath{\mathrm{\tau}}}}1
{φ}{{\ensuremath{\mathrm{\varphi}}}}1
{χ}{{\ensuremath{\mathrm{\chi}}}}1
{ψ}{{\ensuremath{\mathrm{\psi}}}}1
{ω}{{\ensuremath{\mathrm{\omega}}}}1
{Γ}{{\ensuremath{\mathrm{\Gamma}}}}1
{Δ}{{\ensuremath{\mathrm{\Delta}}}}1
{Θ}{{\ensuremath{\mathrm{\Theta}}}}1
{Λ}{{\ensuremath{\mathrm{\Lambda}}}}1
{Σ}{{\ensuremath{\mathrm{\Sigma}}}}1
{Φ}{{\ensuremath{\mathrm{\Phi}}}}1
{Ξ}{{\ensuremath{\mathrm{\Xi}}}}1
{Ψ}{{\ensuremath{\mathrm{\Psi}}}}1
{Ω}{{\ensuremath{\mathrm{\Omega}}}}1
{ℵ}{{\ensuremath{\aleph}}}1
{≤}{{\ensuremath{\leq}}}1
{≥}{{\ensuremath{\geq}}}1
{≠}{{\ensuremath{\neq}}}1
{≈}{{\ensuremath{\approx}}}1
{≡}{{\ensuremath{\equiv}}}1
{≃}{{\ensuremath{\simeq}}}1
{≤}{{\ensuremath{\leq}}}1
{≥}{{\ensuremath{\geq}}}1
{∂}{{\ensuremath{\partial}}}1
{∆}{{\ensuremath{\triangle}}}1 %
{∫}{{\ensuremath{\int}}}1
{∑}{{\ensuremath{\mathrm{\Sigma}}}}1
{Π}{{\ensuremath{\mathrm{\Pi}}}}1
{⊥}{{\ensuremath{\perp}}}1
{∞}{{\ensuremath{\infty}}}1
{∂}{{\ensuremath{\partial}}}1
{∓}{{\ensuremath{\mp}}}1
{±}{{\ensuremath{\pm}}}1
{×}{{\ensuremath{\times}}}1
{⊕}{{\ensuremath{\oplus}}}1
{⊗}{{\ensuremath{\otimes}}}1
{⊞}{{\ensuremath{\boxplus}}}1
{∇}{{\ensuremath{\nabla}}}1
{√}{{\ensuremath{\sqrt}}}1
{⬝}{{\ensuremath{\cdot}}}1
{•}{{\ensuremath{\cdot}}}1
{∘}{{\ensuremath{\circ}}}1
{⁻}{{\ensuremath{^{-}}}}1
{▸}{{\ensuremath{\blacktriangleright}}}1
{∧}{{\ensuremath{\wedge}}}1
{∨}{{\ensuremath{\vee}}}1
{¬}{{\ensuremath{\neg}}}1
{⊢}{{\ensuremath{\vdash}}}1
{⟨}{{\ensuremath{\langle}}}1
{⟩}{{\ensuremath{\rangle}}}1
{↦}{{\ensuremath{\mapsto}}}1
{←}{{\ensuremath{\leftarrow}}}1
{<-}{{\ensuremath{\leftarrow}}}1
{→}{{\ensuremath{\rightarrow}}}1
{↔}{{\ensuremath{\leftrightarrow}}}1
{⇒}{{\ensuremath{\Rightarrow}}}1
{⟹}{{\ensuremath{\Longrightarrow}}}1
{⇐}{{\ensuremath{\Leftarrow}}}1
{⟸}{{\ensuremath{\Longleftarrow}}}1
{∩}{{\ensuremath{\cap}}}1
{∪}{{\ensuremath{\cup}}}1
{⊂}{{\ensuremath{\subseteq}}}1
{⊆}{{\ensuremath{\subseteq}}}1
{⊄}{{\ensuremath{\nsubseteq}}}1
{⊈}{{\ensuremath{\nsubseteq}}}1
{⊃}{{\ensuremath{\supseteq}}}1
{⊇}{{\ensuremath{\supseteq}}}1
{⊅}{{\ensuremath{\nsupseteq}}}1
{⊉}{{\ensuremath{\nsupseteq}}}1
{∈}{{\ensuremath{\in}}}1
{∉}{{\ensuremath{\notin}}}1
{∋}{{\ensuremath{\ni}}}1
{∌}{{\ensuremath{\notni}}}1
{∅}{{\ensuremath{\emptyset}}}1
{∖}{{\ensuremath{\setminus}}}1
{†}{{\ensuremath{\dag}}}1
{ℕ}{{\ensuremath{\mathbb{N}}}}1
{ℤ}{{\ensuremath{\mathbb{Z}}}}1
{ℝ}{{\ensuremath{\mathbb{R}}}}1
{ℚ}{{\ensuremath{\mathbb{Q}}}}1
{ℂ}{{\ensuremath{\mathbb{C}}}}1
{⌞}{{\ensuremath{\llcorner}}}1
{⌟}{{\ensuremath{\lrcorner}}}1
{⦃}{{\ensuremath{\{\!|}}}1
{⦄}{{\ensuremath{|\!\}}}}1
{‖}{{\ensuremath{\|}}}1
{₁}{{\ensuremath{_1}}}1
{₂}{{\ensuremath{_2}}}1
{₃}{{\ensuremath{_3}}}1
{₄}{{\ensuremath{_4}}}1
{₅}{{\ensuremath{_5}}}1
{₆}{{\ensuremath{_6}}}1
{₇}{{\ensuremath{_7}}}1
{₈}{{\ensuremath{_8}}}1
{₉}{{\ensuremath{_9}}}1
{₀}{{\ensuremath{_0}}}1
{ᵢ}{{\ensuremath{_i}}}1
{ⱼ}{{\ensuremath{_j}}}1
{ₐ}{{\ensuremath{_a}}}1
{¹}{{\ensuremath{^1}}}1
{ₙ}{{\ensuremath{_n}}}1
{ₘ}{{\ensuremath{_m}}}1
{ₚ}{{\ensuremath{_p}}}1
{↑}{{\ensuremath{\uparrow}}}1
{↓}{{\ensuremath{\downarrow}}}1
{...}{{\ensuremath{\ldots}}}1
{·}{{\ensuremath{\cdot}}}1
{▸}{{\ensuremath{\triangleright}}}1
{Σ}{{\color{symbolcolor}\ensuremath{\Sigma}}}1
{Π}{{\color{symbolcolor}\ensuremath{\Pi}}}1
{∀}{{\color{symbolcolor}\ensuremath{\forall}}}1
{∃}{{\color{symbolcolor}\ensuremath{\exists}}}1
{λ}{{\color{symbolcolor}\ensuremath{\mathrm{\lambda}}}}1
{\$}{{\color{symbolcolor}\$}}1
{:=}{{\color{symbolcolor}:=}}1
{=}{{\color{symbolcolor}=}}1
{<|>}{{\color{symbolcolor}<|>}}1
{<\$>}{{\color{symbolcolor}<\$>}}1
{+}{{\color{symbolcolor}+}}1
{*}{{\color{symbolcolor}*}}1,
morecomment=[s][\color{commentcolor}]{/-}{-/},
morecomment=[l][\itshape \color{commentcolor}]{--},
showstringspaces=false,
keepspaces=true,
morestring=[b]",
morestring=[d],
tabsize=3,
extendedchars=true,
sensitive=true,
breaklines=true,
breakatwhitespace=true,
basicstyle=\ttfamily\small,
captionpos=b,
columns=[l]fullflexible,
identifierstyle={\ttfamily\color{black}},
keywordstyle=[1]{\ttfamily\color{keywordcolor}},
keywordstyle=[2]{\ttfamily\color{sortcolor}},
keywordstyle=[3]{\ttfamily\color{tacticcolor}},
keywordstyle=[4]{\ttfamily\color{attributecolor}},
stringstyle=\ttfamily,
commentstyle={\ttfamily\footnotesize },
}
\av{\lstset{basicstyle={\small\ttfamily},
	aboveskip=\medskipamount, belowskip=\medskipamount}}

\usepackage{booktabs}

\usepackage{todonotes}

\usepackage{graphicx}
\usepackage{caption}
\usepackage{amsfonts}
\usepackage{amsmath}
\usepackage{amsthm}
\newtheorem{example}{Example}
\usepackage{booktabs}
\usepackage{siunitx}
\usepackage{tikz}
\usepackage{pgfplots}
\pgfplotsset{compat=1.18}
\usetikzlibrary{shapes.geometric, shapes.symbols, arrows.meta, positioning, fit, backgrounds}

\av{\usepackage[hidelinks]{hyperref}\urlstyle{rm}}
\usepackage{cleveref}

\cv{\pdfinfo{
/TemplateVersion (2027.1)
}}
\av{\hypersetup{%
  pdftitle={LeanCSP: A Framework for Certifying Constraint Reformulation and Solving in Lean},%
  pdfauthor={Pablo Manrique, Stefan Szeider}}}

\title{LeanCSP: A Framework for Certifying Constraint Reformulation and Solving in Lean}

\cv{
\author{
    Anonymous Submission
}
\affiliations{
}
}
\av{
\author{
    Pablo Manrique \qquad Stefan Szeider\\[4pt]
    \small Algorithms and Complexity Group\\[-2pt]
    \small TU Wien, Vienna, Austria\\[-2pt]
    \small \texttt{\{pmanriqu,sz\}@ac.tuwien.ac.at}
}
}

\begin{document}

\maketitle
\av{\thispagestyle{empty}}

\begin{abstract}

Constraint programming is a core technology for solving complex combinatorial problems in scheduling, planning, configuration, and verification. Trusting its results therefore demands guarantees at two levels: that reformulations applied beforehand are semantics-preserving, and that solvers produce correct answers. In this work, we introduce a framework that addresses both verification levels in the Lean theorem prover: it can be used to prove formulation-level properties, such as equivalence, equisatisfiability, and the correctness of symmetry-breaking constraints, parametrically for entire problem families; and to check solver-produced certificates for individual instances via translation backends to external formats such as MiniZinc, SMT-LIB, and OPB. Combining both levels yields an end-to-end workflow that establishes the satisfiability or unsatisfiability of a constraint problem without trusting the external solver. Experimental results show that our framework's verified symmetry breaking also pays off in practice: a single parametric proof per problem family, reused across all instance sizes, reduces solver search effort by a factor of up to $2\times10^7$, while the entire in-Lean certification stays affordable, taking at most a few minutes for our largest instances.

\end{abstract}

\begin{cvblock}
\begin{links}
\end{links}
\end{cvblock}
\begin{avblock}
\begin{links}
    \link{Code}{https://github.com/leansolving/leancsp}
\end{links}
\end{avblock}

\section{Introduction}
\label{sec:intro}

\begin{figure*}[t]
\centering
\definecolor{nodemodel}{HTML}{B9D7EA}
\definecolor{nodethm}{HTML}{CFE8CF}
\definecolor{nodeproc}{HTML}{F3D7B6}
\definecolor{zonetrust}{HTML}{EEF4FA}
\definecolor{zoneext}{HTML}{F7F1E8}
\definecolor{bordercol}{HTML}{4A4A4A}
\definecolor{textcol}{HTML}{222222}
\av{\begin{adjustbox}{max width=\textwidth}}%
\begin{tikzpicture}[
  font=\footnotesize,
  every node/.append style={text=textcol},
  box/.style={draw=bordercol, rounded corners, fill=nodemodel, align=center, text width=1.8cm, minimum height=0.9cm, inner sep=2pt},
  thm/.style={box, fill=nodethm},
  proc/.style={box, fill=nodeproc},
  ell/.style={draw=bordercol, fill=white, ellipse, align=center, inner sep=1pt, minimum width=1.4cm, minimum height=0.9cm},
  arr/.style={-{Latex[length=1.6mm]}, semithick, draw=bordercol},
  ext/.style={draw=bordercol, dashed, fill=white, align=center, text width=1.4cm, minimum height=0.9cm, inner sep=2pt},
  doc/.style={draw=bordercol, fill=white, align=center, shape=tape, tape bend top=none, inner sep=4pt},
]
\node[box] (pP)    at (0,0)       {CSP $\mathcal{P}$};
\node[box] (pPp)   at (0,-2.3)    {CSP $\mathcal{P}'$};
\node[thm] (equiv) at (-2.6,-1.15){equiv/equisat thm};
\node[thm] (thm)   at (2.5,-1.15) {(un)satisfiable thm};
\node[proc] (decide) at (5.1,0)    {Lean\\\texttt{decide}};
\node[proc] (pblean) at (5.1,-2.3) {PBLean};
\node[ell] (satcert)   at (7.5,0)    {SAT\\cert};
\node[ell] (unsatcert) at (7.5,-2.3) {UNSAT\\cert};
\node[ext] (solver)    at (10.1,0)    {Solver};
\node[doc, text width=2.7cm] (file) at (10.1,-2.3) {\texttt{.mzn/.smt/.opb} file};

\node[anchor=south west, font=\footnotesize, text=purple!55!black] (reformlabel) at ([yshift=8pt]equiv.west |- pP.north) {Verified Reformulation};
\node[anchor=south east, font=\footnotesize, text=teal!65!black] (solvelabel) at ([yshift=8pt]file.east |- pP.north) {Verified Solving};

\begin{scope}[on background layer]
\node[fill=black!5, draw=bordercol, line width=0.8pt, rounded corners,
    fit=(reformlabel)(equiv)(pP)(pPp)(thm)(decide)(pblean), inner sep=9pt] (leanbox) {};
\node[fill=purple, fill opacity=0.12, rounded corners,
      fit=(equiv)(pP)(pPp), inner sep=5pt] (reformbox) {};
\node[fill=teal, fill opacity=0.12, rounded corners,
      fit=(pP)(pPp)(thm)(decide)(pblean)(satcert)(unsatcert)(solver)(file), inner sep=5pt] (solvebox) {};
\end{scope}
\node[anchor=south, font=\footnotesize] at ([yshift=2pt]leanbox.north) {Lean};

\draw[{Latex[length=1.6mm]}-{Latex[length=1.6mm]}, semithick, draw=bordercol] (pP) -- (pPp);
\draw[arr] (equiv.east) -- (0,-1.15);

\draw[arr] (decide.west) -- (thm.east);
\draw[arr] (pblean.west) -- (thm.east);

\draw[arr] (thm.west) -- (pP.east);
\draw[arr] (thm.west) -- (pPp.east);

\draw[arr] (satcert.west)   -- (decide.east);
\draw[arr] (unsatcert.west) -- (pblean.east);

\draw[arr] (solver.west)  -- (satcert.east);
\draw[arr] (solver.west) -- (unsatcert.east);

\draw[arr] (file.north) -- (solver.south);

\draw[arr] (pPp.south) -- ++(0,-0.55) -| (file.south)
      node[pos=0.25, below, font=\scriptsize]{translation};
\end{tikzpicture}%
\av{\end{adjustbox}}
\caption{Overview of LeanCSP. A CSP $\mathcal{P}$ can be formulated in Lean, and a reformulation (an equivalence or equisatisfiability theorem) that yields another CSP $\mathcal{P'}$ can be proven correct at the parametric level. At the instance level, a CSP is translated to an external solver format; the solver's SAT or UNSAT certificate is checked back in Lean (by \texttt{decide} or PBLean, respectively), establishing an (un)satisfiability theorem. The external solvers are outside the trusted Lean base: every certificate produced is checked back against the original problem inside Lean, so any possible errors or bugs in the solvers cannot produce a false guarantee, only a check that fails.}
\label{fig:overview}
\end{figure*}
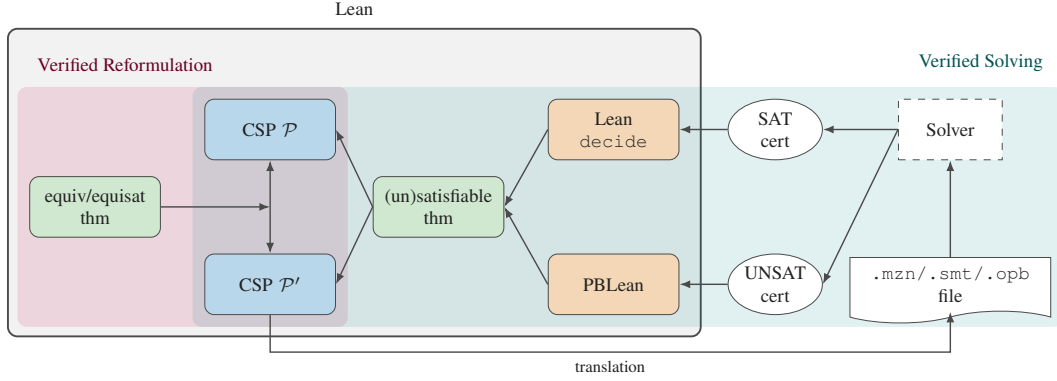

Constraint satisfaction problems (CSPs) constitute a powerful framework for modeling and solving a wide range of combinatorial problems, from scheduling and resource allocation to hardware verification and combinatorial design~\cite{tsangFoundationsConstraintSatisfaction1993, dechterConstraintProcessing2003a,rossiHandbookConstraintProgramming2007}. Operationally, practitioners often reformulate CSPs to improve solver performance, switching between alternative encodings or adding symmetry-breaking constraints~\cite{gentSymmetryConstraintProgramming2006}.  
Moreover, different symmetry-breaking constraints can be added to the problem, with the intent of pruning symmetric solutions and speeding up the solving process. Ensuring that these modifications do not contain errors that can silently eliminate solutions or introduce spurious ones is essential for correctness.
Crucially, such an error can be invisible to testing: we exhibit a plausible symmetry-breaking constraint for the Schur problem that preserves satisfiability on every instance up to $n=12$ yet silently flips the answer exactly at $n=13$, where a solver then certifies a false bound on the third Schur number $S(3)$. Our proposed framework rejects this constraint, because the required symmetry proof cannot be given. 
Furthermore, once the problem is formulated as a CSP, solvers are used to find solutions that satisfy the specified constraints. However, modern solvers are complex and constantly evolving software artifacts that might contain bugs~\cite{vanrooseMutationalFuzzTesting2024,mansurDetectingCriticalBugs2020}, so checking that the answer given by the solver is correct is another important step to obtain formal guarantees.

These two formal verification problems are addressed in the literature using two main approaches: \emph{proof assistants} and \emph{proof logging}. Proof assistants (or \emph{interactive theorem provers}, ITPs), like Rocq~\cite{therocqdevelopmentteamRocqProver2025}, Isabelle/HOL~\cite{nipkowIsabelleHOLProof2002}, Agda~\cite{stumpVerifiedFunctionalProgramming2016} or Lean~\cite{mouraLean4Theorem2021}, are systems that let users develop formal proofs validated by a small trusted kernel built on a foundational logic. Their expressivity allows one to obtain general guarantees independent of the instance size at the formulation level: related work is verifying SAT encodings, proving that they are equivalent to the mathematical properties that they represent~\cite{cruz-filipeFormallyVerifyingSolution2019,abdulazizFormallyVerifiedSATBased2023,codelVerifiedEncodingsSAT2023,subercaseauxFormalVerificationEmpty2024}; certified constraint solvers~\cite{carlierCertifiedConstraintSolver2012,colibrics}; and, closest to our work, proving that a transformation of CSPs preserves satisfiability~\cite{duboisFormallyVerifiedTransformation2020}.
In proof logging, solvers emit certificates of their answers that small separate checkers validate per instance, so one proof is obtained per solver run. This is a standard practice in SAT~\cite{heuleTrimmingCheckingClausal2013,wetzlerDRATtrimEfficientChecking2014,cruz-filipeEfficientCertifiedRAT2017} and has been extended to SMT~\cite{barrettProofsinSatisfiabilityModuloTheories}, or CP~\cite{gochtAuditableConstraintProgramming2022,flippoMultiStageProofLogging2024}. Proof logging can also be used for certifying reformulations at the instance level, as VeriPB~\cite{gochtCertifyingParityReasoning2021} can certify dominance and symmetry breaking~\cite{bogaertsCertifiedDominanceSymmetry2023}. Checkers are formally verified in some cases~\cite{gochtEndtoEndVerificationSubgraph2024}, and in Lean, \texttt{bv\_decide}~\cite{bovingInteractiveBitvectorReasoning2025} can be used to check LRAT proofs, and PBLean~\cite{szeiderPBLeanPseudoBooleanProof2026} does the same for pseudo-Boolean certificates. These last two approaches rely on reflection: a checker proved sound once in Lean is run as compiled code, and a successful check yields a reusable Lean theorem rather than a mere external verdict.

Our contribution is to tackle both verification problems using the Lean~4 theorem prover~\cite{mouraLean4Theorem2021}, based on dependent types and the calculus of inductive constructions, and whose extensive \texttt{mathlib}~\cite{LeanMathematicalLibrary2020} library and a growing ecosystem of development tools~\cite{mohamedLeanSMTSMTTactic2025,LeanbackMCPServer} has enabled landmark formalizations, including the resolution of Marton's conjecture~\cite{gowersConjectureMarton2025}, the sphere eversion~\cite{vandoornFormalisingHPrincipleSphere2023}, and a verified typechecker for Lean itself~\cite{carneiroLean4LeanVerifyingTypechecker2025}.
We develop a formalization of constraint satisfaction problems in Lean that allows us to specify constraint problems using a library of over 40 different constraints, and to reason about them in a general way. In particular, we formalize properties such as \emph{equivalence} (the preservation of all solutions of a problem) and \emph{equisatisfiability} (the preservation of a problem's satisfiability), the latter capturing symmetry-breaking constraints as a special case. 
The expressivity of Lean allows us to prove these properties once for entire parameterized families of CSPs---e.g., for all $n$ in the $n$-queens problem---so instances of all possible sizes automatically inherit them.
Our framework, named LeanCSP and illustrated in \Cref{fig:overview}, also enables the automatic translation of CSPs formulated in Lean into different formats once instantiated, so that external solvers can solve them. We include translation backends for MiniZinc~\cite{nethercoteMiniZincStandardCP2007}, and the SMT-LIB~\cite{barrettSMT-LIB} and OPB~\cite{rousselInputOutputFormat} formats, allowing us to use constraint, SMT, and pseudo-Boolean solvers, respectively. Satisfiability certificates (i.e., solutions) are reimported into Lean after the solver run finishes and are checked against the original CSP to certify satisfaction. In the case of unsatisfiable problems, which we will focus on in greater detail in this paper, the pseudo-Boolean solver \textit{RoundingSat}~\cite{elffersDivideConquer27th2018} is used to produce VeriPB~\cite{bogaertsCertifiedDominanceSymmetry2023} proofs that can be independently checked inside Lean via PBLean~\cite{szeiderPBLeanPseudoBooleanProof2026}.
Although a SAT path is equally feasible using \texttt{bv\_decide}, we follow this pseudo-Boolean path, which our preliminary experiments (\Cref{sec:experiments}) found better suited to the problems we target through its cutting planes reasoning. 
Since PBLean can only check the unsatisfiability of a pseudo-Boolean model inside Lean, we prove, for every constraint in our library, that its pseudo-Boolean representation in Lean is satisfied whenever the original CSP constraint is. This translation soundness theorem lets us certify the unsatisfiability of any CSP expressible in our library.
This way, and to the best of our knowledge, we establish the first formally verified end-to-end pipeline for constraint solving at the level of a general constraint language: a CSP formulated in Lean and taken as the specification of the problem is proved satisfiable or unsatisfiable by delegating the search to external solvers whose output is checked back against the CSP itself, so that neither the translation nor the solvers need to be trusted, and uniformly for every CSP built from our library of constraints. This sets us apart from proof-logging approaches,
whose certificates, even when validated by a formally verified checker, concern the low-level formula the solver works on rather than yielding an unsatisfiability theorem about the CSP itself, and from verified solvers built inside a proof assistant, which are correct by construction at the expense of performance.
Furthermore, combining both the reformulation correctness and the certificate checking layers lets us prove the (un)satisfiability of CSPs that have not been directly solved themselves, by first proving equisatisfiability with an optimized constraint formulation that may use symmetry-breaking constraints and other solving speedup techniques, and then solving this second problem via an external solver and checking its output. We apply this pipeline to certify concrete combinatorial results in Lean, for instance the Schur number $S(4) = 44$ and the bounds $R(3,3) \leq 6$ and $W(2,3) \leq 9$. \cv{The complete Lean development is included in the supplementary material.}%
\av{The complete Lean development is available at \url{https://github.com/leansolving/leancsp}.}

\section{Formalization}
\label{sec:formalization}

In this section, we will describe our formalization of constraint satisfaction problems in Lean, as well as the definitions of the central reformulation properties that we consider. We assume only a passing familiarity with Lean, and we will provide high-level explanations of the different concepts. However, we include some code snippets from the actual Lean formalization that can help illustrate those concepts. We refer to the original documentation~\cite{LeanLanguageReference} for further details about Lean's syntax.
\av{
Our implementation is publicly available at
\begin{center}
\url{https://github.com/leansolving/leancsp}.
\end{center}
Code listings in this paper are lightly edited for presentation; the exact statements can be found in the repository.}

\subsection{Basic Definitions}
\label{sec:basicdefinitions}

Following the standard definition~\cite{rossiHandbookConstraintProgramming2007}, a CSP $\mathcal{P} = (X, D, C)$ comprises variables $X = (x_1, \ldots, x_n)$, domains $D = (D_1, \ldots, D_n)$ with $x_i \in D_i$, and constraints $C = (C_1, \ldots, C_m)$ where each $C_j$ is a pair $(S_j, R_j)$ of a scope $S_j \subseteq X$ and a relation $R_j$ over the domains of variables in $S_j$.
Since Lean is based on dependent type theory, it is more natural to define CSPs in Lean using dependent types rather than sets and tuples as in the standard CP definition above. More precisely, given any type representing the variables \lstinline|VarIndex|, and a function \lstinline|DomainType| mapping each variable to its (arbitrary) domain type, we define a CSP as a list of constraints:

\begin{lstlisting}
abbrev CSP (VarIndex : Type) (DomainType : VarIndex → Type) :=
  List (DynamicConstraint VarIndex DomainType)
\end{lstlisting}

Since all elements of a list must have the same type, we use the \lstinline|DynamicConstraint| wrapper type that allows expressing constraints without fixing their arity. The underlying \lstinline|Constraint| type is defined by a scope and a checker function that expresses the relation:

\begin{lstlisting}
structure Constraint (VarIndex : Type) (DomainType : VarIndex → Type) 
    (n : ℕ) where
  scope : Vector VarIndex n
  check : ScopeValues VarIndex DomainType scope → Bool
\end{lstlisting}

Here, \lstinline|ScopeValues| is a dependent function that maps each scope index to the corresponding domain type. Intuitively, it just represents the product $\prod_{x_i \in S_j} D_i$.

Other basic concepts regarding CSPs are also formalized: an assignment is a function that maps variables to values in their respective domains; a solution of a CSP is an assignment that satisfies all constraints of the problem; and a CSP is satisfiable if it has a solution.

\begin{example}[Schur triples]
    \label{ex:schur}
    Consider the integers $\{1, \ldots, n\}$. Can they be partitioned into $k$ classes such that no class contains numbers $a$, $b$, $c$ (not necessarily distinct) with $a + b = c$? We can formulate this problem as a CSP $\mathcal{P}$ using $n$ integer variables $x_1, \ldots, x_n$, one per number, each ranging over the $k$ possible classes (or ``colors'') $\{0, \ldots, k-1\}$. Then, for every triple $(a, b, c)$ with $a + b = c$, we use a constraint stating that $x_a$, $x_b$, and $x_c$ are not all equal. The largest $n$ for which this problem is satisfiable is the Schur number $S(k)$~\cite{heuleSchurNumberFive2018}; for four colors, $S(4) = 44$, so the instance with $k = 4$ is satisfiable for $n \leq 44$ and unsatisfiable for $n \geq 45$.
\end{example}

\subsection{CSP Equivalence and Equisatisfiability}
\label{equivalence}

Defining precisely when two CSPs are \textit{equivalent}, or when a CSP \textit{encodes} a given problem, is not trivial~\cite{rossiHandbookConstraintProgramming2007}. Taking equivalence to mean having the same set of solutions would make a mere renaming of the variables produce a non-equivalent CSP, although the \textit{information} the problems encode remains the same. We thus use a different definition of equivalence: two CSPs are equivalent if there exists a bijection between their solution sets. In Lean, the definition is the following, where \lstinline|sol_set| is the set of solutions of a CSP:

\begin{lstlisting}
def equivalent (csp₁ : CSP VarIndex₁ DomainType₁) (csp₂ : CSP VarIndex₂ DomainType₂) : Prop :=
  ∃ f : sol_set csp₁ → sol_set csp₂, Function.Bijective f
\end{lstlisting}

Note that this definition is informative only for satisfiable CSPs: any two unsatisfiable CSPs are trivially equivalent, since their empty solution sets are in bijection. 
The notion becomes meaningful through the characterization of equivalence introduced next, which exhibits an explicit correspondence between the solutions of the two problems rather than a mere counting argument.
Given two CSPs $\mathcal{P}$ and $\mathcal{P}'$, and a projection $\pi$ that maps assignments of $\mathcal{P}'$ to assignments of $\mathcal{P}$, we say that $\mathcal{P}$ and $\mathcal{P}'$ are $\pi$-equivalent if:

\begin{enumerate}
  \item Solutions of $\mathcal{P}'$ map to solutions of $\mathcal{P}$,
  \item Every solution of $\mathcal{P}$ has a preimage, and
  \item $\pi$ is injective on solutions.
\end{enumerate}

If we prove that two CSPs $\mathcal{P}$ and $\mathcal{P}'$ are $\pi$-equivalent, then they are equivalent via the bijection $\pi|_{\mathrm{sol}(\mathcal{P}')}$. The $\pi$-equivalence definition is slightly more convenient to use when reasoning on concrete CSPs, since it is usually easier to define a map between the assignment sets explicitly than between the solution sets, where constraint satisfaction is implicit.

Under the natural assumption that the assignment type of $\mathcal{P}$ is non-empty, i.e., every variable has at least one domain value, we prove that two CSPs are equivalent if and only if they are $\pi$-equivalent.

\begin{lstlisting}
theorem equivalent_iff_pi_equivalent [Nonempty (Assignment VarIndex₁ DomainType₁)]
    (csp₁ : CSP VarIndex₁ DomainType₁) (csp₂ : CSP VarIndex₂ DomainType₂) :
  equivalent csp₂ csp₁ ↔ 
    ∃ π, pi_equivalent csp₁ csp₂ π
\end{lstlisting}
 
On the other hand, the definition of equisatisfiability that we use is the natural one: two CSPs are equisatisfiable if both are satisfiable or both are unsatisfiable.
If two CSPs are equivalent, in particular they are equisatisfiable.

\begin{example}[Schur triples, continued]
    \label{ex:schurequiv}
    An alternative formulation $\mathcal{P}'$ of the Schur problem of \cref{ex:schur} uses $n \cdot k$ Boolean variables $y_{i,d}$, one per number $i$ and color $d$ (a one-hot encoding): a cardinality constraint $\sum_{d} y_{i,d} = 1$ states that each number receives exactly one color, and for every triple $(a, b, c)$ with $a + b = c$ and every color $d$, the constraint $y_{a,d} + y_{b,d} + y_{c,d} \leq 2$ forbids the monochromatic triple. The projection $\pi$ maps an assignment of $\mathcal{P}'$ to the assignment of $\mathcal{P}$ that gives each number $i$ the unique color $d$ with $y_{i,d} = 1$. We prove in Lean that $\mathcal{P}$ and $\mathcal{P}'$ are $\pi$-equivalent, parametrically for all $n$ and $k$, so equivalence follows by the characterization above.
\end{example}

\subsection{Symmetry}
\label{sec:symmetry}

Symmetry in CSPs has been widely studied, and many definitions have appeared in the literature~\cite{cohenSymmetryDefinitionsConstraint2005}. We define a \textit{symmetry} as a permutation on the variables or the domain sets that preserves solutions, and a \textit{symmetry-breaking constraint} as a constraint such that, given a solution of the original problem, there exists a symmetry that transforms it into a solution of the extended CSP that includes the new constraint.

We distinguish between \emph{variable symmetries}, which are permutations of the variable indices, and \emph{domain symmetries}, which are permutations of the values in the domains. A variable symmetry $\beta$ is a bijection on the variable index set such that for any solution, permuting the variables according to $\beta$ yields another solution.
Note that, since in general the value domains may be different for each variable,
it is required that, for each variable $v$, the domain type of $v$ is the same as the domain type of $\beta(v)$.
Domain symmetries are defined similarly, but instead of permuting variable indices, they permute the values in the domains. Since, in the case of domains, they are not necessarily the same for all variables, we need to provide permutations for the domain of each variable. In Lean, we do this by defining a domain equivalence (or permutation) family: a dependent function that assigns each variable a permutation of its domain type.
In order to apply a domain permutation family $\delta$ to an assignment $\alpha$, we need to map each variable $v$ to the value $\delta(v)(\alpha(v))$ (this is what the \lstinline|apply| function does in Lean). A domain symmetry is then a domain equivalence family $\delta$ such that for any solution, applying $\delta$ to it yields another solution.

\begin{lstlisting}
def DomainSymmetry (csp : CSP VarIndex DomainType) 
    (δ : DomainEquivFamily VarIndex DomainType) : Prop :=
  ∀ assignment, is_solution csp assignment → is_solution csp (δ.apply assignment)
\end{lstlisting}

Once the variable and domain symmetries on CSPs are defined, we can formalize the symmetry-breaking constraints (SBCs) described at the start of this subsection. In Lean, we define domain symmetry-breaking constraints in the following way (the variable case is analogous):

\begin{lstlisting}
def domainSymmetryBreakingConstraint (csp : CSP VarIndex DomainType) 
    (c : DynamicConstraint VarIndex DomainType) : Prop :=
  ∀ assignment : Assignment VarIndex DomainType, is_solution csp assignment →
    ∃ δ : DomainEquivFamily VarIndex DomainType, DomainSymmetry csp δ ∧ is_solution (extended_csp csp c) (δ.apply assignment)
\end{lstlisting}

Finally, these definitions allow us to prove that adding a symmetry-breaking constraint to a CSP preserves satisfiability, since the underlying symmetry transforms solutions of the original problem into solutions of the extended problem.

\begin{example}[Schur triples, continued]
    \label{ex:schursb}
    In the Schur CSP $\mathcal{P}$ of \Cref{ex:schur}, the $k$ colors are interchangeable: permuting the colors $\{0, \ldots, k-1\}$ in any solution yields another solution, since the not-all-equal triple constraints depend only on which numbers share a color, not on the color names. Every such color permutation is therefore a domain symmetry of $\mathcal{P}$, and the value-precedence constraint of the next subsection breaks the whole symmetry group at once by forcing the colors to first appear in increasing order, keeping a single representative of each class. Discharging this obligation, that the permutation really maps solutions to solutions, is exactly the step that the cautionary example of \Cref{sec:cautionary} cannot complete for its reversal candidate, which is not a symmetry of the Schur problem.
\end{example}

\subsection{Examples and Proof Effort}
\label{sec:examplesandeffort}

To demonstrate the applicability of our formalization, we model several standard problem families as parameterized CSPs in Lean and prove reformulation properties about them. These problems are formulated using a library of more than 40 predefined constraints, organized into five families: comparison constraints (\texttt{eq}, \texttt{ne}, \texttt{lt}, \texttt{le}, \ldots), arithmetic constraints (linear combinations, products, absolute values), global constraints (\texttt{alldifferent}, cardinality, \texttt{element}), Boolean gates and logical constraints such as implication and equivalence, and bound constraints that restrict the value ranges of the variables. 

We prove two kinds of results. For six families we provide two different CSP formulations of the same problem and prove them equivalent through the $\pi$-equivalence characterization of \Cref{equivalence}: $n$-queens, mutilated chessboard, Latin squares, graph coloring, Schur triples, and Sudoku (we consider classical constraint formulations with several possible integer values versus one-hot formulations with binary values). 
For these problem families and five more (perfect matchings, Langford, pigeonhole principle, and Ramsey and van der Waerden numbers), we define symmetry-breaking constraints and prove them sound, thereby obtaining an equisatisfiability result between the base CSP and its extension. One of them deserves special mention: \emph{value precedence}---already mentioned in \Cref{ex:schursb}---an SBC that targets problems whose values are interchangeable. The idea is to force the values to \emph{first appear} in a fixed order, thus keeping a single representative of each value symmetry class. We prove this constraint sound once, generically and parametrically in the number of interchangeable values, and then reuse it across five families through short per-family glue lemmas that exhibit the relevant value permutation as a symmetry.

All these proofs are parametric in the instance size, so each property is discharged once and automatically inherited by every instance of the family. \Cref{tab:proofeffort-loc} reports the size of the resulting development, in lines of Lean code, broken down by family and by kind of result. In total, the equivalence proofs comprise 6,183 lines, the non-value-precedence symmetry-breaking soundness proofs 2,349 lines, and the value-precedence glue 380 lines; the latter apply the generic value-precedence proof (149 lines) established once and for all. The general formalization, which includes the basic definitions and reformulation properties, and the general results that relate symmetry, equisatisfiability and equivalence, sum up to about 2000 lines of Lean code.

\begin{table}[t]
\centering\small
\begin{tabular}{l r r r}
\toprule
Family & Equiv. & SB & VP \\
\midrule
$n$-queens        & 1,791 & 367 & --- \\
Mutilated         & 1,216 & 323 & --- \\
Latin square      &   978 & 486 & --- \\
Graph coloring   &   791 & 254 &  71 \\
Schur             &   754 & 228 &  91 \\
Sudoku            &   653 & 203 & --- \\
Matching          &   --- & 275 & --- \\
Langford          &   --- & 213 & --- \\
Pigeonhole        &   --- & --- &  90 \\
Ramsey            &   --- & --- &  64 \\
Van der Waerden   &   --- & --- &  64 \\
\midrule
\multicolumn{3}{l}{Generic value-precedence proof} & 149 \\
\midrule
Total             & 6,183 & 2,349 & 529 \\
\bottomrule
\end{tabular}
\caption{Proof effort, in lines of Lean code, per problem family: equivalence proof (Equiv.), symmetry-breaking soundness proof in general (SB), and value-precedence (VP). The per-family VP entries are glue that instantiates a single generic value-precedence proof (149 lines), established once for any CSP and any number of interchangeable values. For Schur and graph coloring, we provide a weaker symmetry constraint that only fixes the first color, alongside VP. A dash means no such result was developed for that family.}
\label{tab:proofeffort-loc}
\end{table}

\section{Certificate Generation and Checking}
\label{sec:cert}

Our framework supports the translation of CSPs to external solver formats for a subclass of CSPs that is expressive enough to formulate a wide variety of combinatorial problems. We require the variables to range over finite types, their values to be integers, and the constraints to come from the library of predefined constraints introduced in \Cref{sec:examplesandeffort}. Any CSP following this structure can be translated to the supported external formats once instantiated.

We distinguish two kinds of translation. For MiniZinc and SMT-LIB, each predefined constraint of our Lean library is mapped to a string, in the corresponding output format, that represents that constraint. In the case of the OPB translation, we include an extra step to support unsatisfiability certification: since PBLean can only check proofs against pseudo-Boolean formulas, we first encode to an intermediate PB model inside Lean. This CSP to PB translation is verified to be sound: we prove in Lean that whenever a CSP built from constraints in our library is satisfiable, the corresponding PB model is satisfiable as well. Then, as in the case of MiniZinc and SMT-LIB, we can generate an OPB file from the intermediate PB model. These mappings to the external formats are not formally verified, but they do not need to be trusted: once the solver ran on an instance and produced either a satisfiability (a solution) or an unsatisfiability (a proof) certificate, it can be reimported back into Lean and checked against the original CSP, by directly verifying that the assignment satisfies all constraints in the SAT case, or in the UNSAT case by using PBLean to check a VeriPB proof generated by the solver against the intermediate PB representation, which implies that the original CSP is unsatisfiable by the CSP to PB soundness result. A successful check yields an end-to-end (un)satisfiability result inside Lean, so a faulty translation or an incorrect certificate can only make the check fail, never certify a wrong answer.

In both pipelines, then, the translation steps and the external solvers lie outside the trusted base. What remains trusted is the Lean kernel, which checks every theorem in our library, including the soundness of the PB translation, together with the Lean compiler: PBLean is proved sound in Lean once and then run by reflection as compiled code, so each certified theorem adds to Lean's standard axioms only the two reflection axioms \texttt{Lean.ofReduceBool} and \texttt{Lean.trustCompiler}, which admit the compiled checker's verdict into the kernel and thereby place the Lean compiler in the trusted base, exactly as the \texttt{bv\_decide} tactic does~\cite{bovingInteractiveBitvectorReasoning2025}. Note that PBLean checks the VeriPB proof against the in-Lean PB model rather than against the emitted OPB file, so the OPB serialization also lies outside the trusted base. The CSP itself is also trusted, in the sense that we take it directly as the specification of the problem, its ``mathematical definition'', rather than as an encoding of some separate ground truth.

\begin{example}[Schur triples, continued]
    \label{ex:schurcert}
    Applying both pipelines together lets us certify Schur numbers. On the value-precedence formulation of \Cref{ex:schursb} with $k = 4$ colors, the satisfiability pipeline checks a solution at $n = 44$ while the unsatisfiability pipeline checks a VeriPB proof that $n = 45$ is already too many; its equisatisfiability with the plain Schur CSP carries both verdicts over, establishing $S(4) = 44$ in Lean. The SAT witnesses were produced via MiniZinc (Gecode~\cite{schulteModelingProgrammingGecode} for $S(2)$ and $S(3)$, Chuffed~\cite{chuffed} for $S(4)$) and the UNSAT certificates by RoundingSat~\cite{elffersDivideConquer27th2018}.
    \av{The ${\sim}98$\,MB UNSAT certificate for $n = 45$ is too large to ship with the repository; it can be regenerated with \texttt{experiments/run.py schur}, after which Lean re-checks it.}
\end{example}

\av{\vspace{-\topsep}}%
\begin{lstlisting}
theorem schur_4_exact :
  (schur_csp 44 4).isSatisfiableInt ∧
  ¬ (schur_csp 45 4).isSatisfiableInt
\end{lstlisting}

\subsection{Cautionary Example: An Unsound Reformulation}
\label{sec:cautionary}

The reversal mapping $i \mapsto n+1-i$ is a genuine variable symmetry of the van der Waerden problem and of the modular variant of the Schur problem, so it is tempting to reuse a reversal-based symmetry-breaking constraint, the strict lexicographic leader $x <_{\mathrm{lex}} \mathrm{rev}(x)$, on the ordinary Schur problem of \Cref{ex:schur}. The reformulation survives some tests one would run: for $k = 3$ it stays satisfiable for all $n \leq 12$ ($45$ solutions still satisfy the constraint at $n = 12$), and it first fails at $n = 13 = S(3)$, whose solutions are all palindromes and are therefore all deleted by the strict constraint. If we considered this reformulated CSP as the specification of the problem, we could check an unsatisfiability certificate and falsely ``prove'' that $S(3) < 13$.

Our framework refuses this reformulation. Adding the leader as a sound symmetry-breaking constraint (\Cref{sec:symmetry}) requires two facts: (\emph{i})~the reversal is a symmetry of the Schur CSP, for every $n$; and (\emph{ii})~the leader breaks that symmetry, meaning every solution can be mapped by the symmetry to one that satisfies the leader. Both are false. Fact (\emph{i}) already fails at $n = 3$, as one checks by hand: the solution $[0,1,1]$ reverses to $[1,1,0]$, which colors the numbers $1$, $1$, and $2$ alike although $1 + 1 = 2$.

\begin{lstlisting}
theorem reversal_not_schur_symmetry :
    ¬ VariableSymmetry (schur_csp 3 3) (reversal 3)
\end{lstlisting}

This disproof at $n = 3$ already refutes the family-level obligation that the reversal is a symmetry for all $n$ (the same witness pattern applies for every $3 \leq n \leq 12$). The leader is thus not an SBC: there is no symmetry to break in the first place.
This missing symmetry is not, however, the real cause of the false result at $n = 13$: for this specific value of $n$, the reversal actually is a symmetry, since every solution is a palindrome and is fixed by it. What fails at $n = 13$ is fact (\emph{ii}). Keeping the SBC sound would require mapping each solution, by the symmetry, onto one that satisfies $x <_{\mathrm{lex}} \mathrm{rev}(x)$; but a palindrome equals its own reversal, so it never satisfies the strict inequality, and the symmetry sends it only to itself. No solution survives the leader, so the extended problem is unsatisfiable.

Note that the framework does not merely \emph{detect} the flaw after the fact; it \emph{refuses} to add the constraint at all, because neither obligation can be proved. This is not to say that proof-logging methods would miss the error: a fully proof-logged reformulation~\cite{bogaertsCertifiedDominanceSymmetry2023} would reject the constraint instance by instance. The vulnerable case is the common workflow, in which a symmetry-breaking constraint is simply added to the model and only the final answer is certified, so that an unsound reformulation silently certifies the wrong result.

\section{Experimental Evaluation}
\label{sec:experiments}

We evaluate, on unsatisfiable instances, three aspects of our pipeline: the scalability of the pseudo-Boolean certification path against a SAT-based alternative, the effect of the verified symmetry-breaking constraints of \Cref{sec:symmetry} on solving, and also the cost of our end-to-end pipeline. All experiments ran on an AMD Ryzen~5 PRO 8540U and 32GB of memory under Ubuntu 24.04, with a timeout of \SI{600}{\second} per solver/checker invocation; each wall-clock time measurement was taken as the median of three runs (deterministic operation counts are run-invariant).

As LeanCSP is a general framework, the purpose of these experiments is not to obtain formal proofs of open problems, but to discuss our design decisions and the feasibility of our methodology.

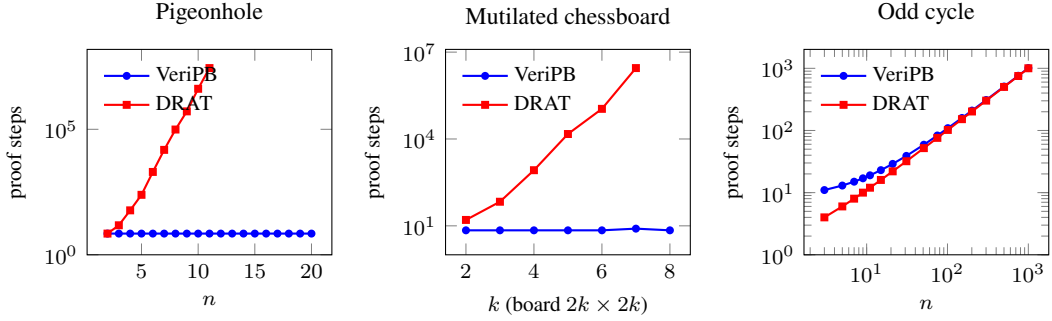
\begin{figure*}[!t]
\centering
\begin{tikzpicture}[baseline=(current axis.south)]
\begin{axis}[width=0.345\textwidth, height=4.3cm, ymode=log, ylabel={proof steps}, ymin=1, tick label style={font=\footnotesize}, label style={font=\footnotesize}, title style={font=\small}, legend style={font=\footnotesize, draw=none, fill=none}, legend cell align=left, legend pos=north west, title={Pigeonhole}, xlabel={$n$}]
\addplot[blue, mark=*, mark size=1.1pt, thick] coordinates {(2,7) (3,7) (4,7) (5,7) (6,7) (7,7) (8,7) (9,7) (10,7) (11,7) (12,7) (13,7) (14,7) (15,7) (16,7) (17,7) (18,7) (19,7) (20,7)};
\addplot[red, mark=square*, mark size=1.1pt, thick] coordinates {(2,7) (3,15) (4,59) (5,245) (6,2000) (7,15143) (8,97904) (9,517775) (10,4113241) (11,27490336)};
\legend{VeriPB, DRAT}
\end{axis}
\end{tikzpicture}\hfill
\begin{tikzpicture}[baseline=(current axis.south)]
\begin{axis}[width=0.345\textwidth, height=4.3cm, ymode=log, ylabel={proof steps}, ymin=1, tick label style={font=\footnotesize}, label style={font=\footnotesize}, title style={font=\small}, legend style={font=\footnotesize, draw=none, fill=none}, legend cell align=left, legend pos=north west, title={Mutilated chessboard}, xlabel={$k$ (board $2k\times2k$)}]
\addplot[blue, mark=*, mark size=1.1pt, thick] coordinates {(2,7) (3,7) (4,7) (5,7) (6,7) (7,8) (8,7)};
\addplot[red, mark=square*, mark size=1.1pt, thick] coordinates {(2,16) (3,68) (4,835) (5,14847) (6,109813) (7,2811941)};
\legend{VeriPB, DRAT}
\end{axis}
\end{tikzpicture}\hfill
\begin{tikzpicture}[baseline=(current axis.south)]
\begin{axis}[width=0.345\textwidth, height=4.3cm, ymode=log, ylabel={proof steps}, ymin=1, tick label style={font=\footnotesize}, label style={font=\footnotesize}, title style={font=\small}, legend style={font=\footnotesize, draw=none, fill=none}, legend cell align=left, legend pos=north west, title={Odd cycle}, xlabel={$n$}, xmode=log]
\addplot[blue, mark=*, mark size=1.1pt, thick] coordinates {(3,11) (5,13) (7,15) (9,17) (11,19) (15,23) (21,29) (31,39) (51,59) (75,83) (101,109) (151,159) (201,209) (301,309) (501,509) (751,759) (1001,1009)};
\addplot[red, mark=square*, mark size=1.1pt, thick] coordinates {(3,4) (5,6) (7,8) (9,10) (11,12) (15,16) (21,22) (31,32) (51,52) (75,76) (101,102) (151,152) (201,202) (301,302) (501,502) (751,752) (1001,1002)};
\legend{VeriPB, DRAT}
\end{axis}
\end{tikzpicture}
\caption{Proof length (proof steps, log scale) versus instance size for the pseudo-Boolean (VeriPB) and SAT (DRAT) pipelines.}
\label{fig:scaling}
\end{figure*}
\begin{table*}[!t]
\centering
\setlength{\tabcolsep}{4pt}
\resizebox{\ifdim\width>\textwidth \textwidth\else\width\fi}{!}{%
\begin{tabular}{l l r r r r r r r r r r r r}
\toprule
 &  & \multicolumn{2}{c}{w/o SBC (lg.)} & \multicolumn{2}{c}{w/ SBC (lg.)} & \multicolumn{2}{c}{Speedup (lg.)} & \multicolumn{2}{c}{Speedup (geo.)} & \multicolumn{2}{c}{Check (native, s)} & \multicolumn{2}{c}{Check (in-Lean, s)} \\
\cmidrule(lr){3-4}\cmidrule(lr){5-6}\cmidrule(lr){7-8}\cmidrule(lr){9-10}\cmidrule(lr){11-12}\cmidrule(lr){13-14}
Family & SBC & det & wall\,(s) & det & wall\,(s) & det & wall & det & wall & w/o & w/ & w/o & w/ \\
\midrule
Clique & vp & $1.1{\times}10^{10}$ & 186 & $5.5{\times}10^{2}$ & 0.0062 & \textbf{19,863,281} & \textbf{30,049} & \textbf{11,043} & \textbf{42} & 183 & 0.00503 & 197 & 1.95 \\
Myciel. & vp & $4.4{\times}10^{9}$ & 63.8 & $1.4{\times}10^{7}$ & 0.71 & \textbf{303} & \textbf{90} & \textbf{48} & \textbf{6.6} & 131 & 1.35 & 145 & 8.14 \\
Schur & vp & \emph{t/o} & \emph{t/o} & $1.0{\times}10^{9}$ & 42.9 & --- & --- & \textbf{7.5} & 0.83 & --- & 47.7 & --- & 67.7 \\
Odd cyc. & $x_0{=}0$ & $4.6{\times}10^{2}$ & 0.0034 & $5.2{\times}10^{1}$ & 0.003 & \textbf{8.8} & \textbf{1.1} & \textbf{8.6} & \textbf{1.0} & 0.00081 & 0.000606 & 5.29 & 5.41 \\
Ramsey & $x_0{=}0$ & $1.2{\times}10^{3}$ & 0.0037 & $6.8{\times}10^{2}$ & 0.0036 & \textbf{1.8} & \textbf{1.0} & \textbf{1.9} & 0.96 & 0.00109 & 0.00105 & 4.29 & 4.29 \\
vdW & vp & $4.3{\times}10^{6}$ & 0.121 & $3.0{\times}10^{6}$ & 0.155 & \textbf{1.4} & 0.78 & \textbf{2.1} & 0.87 & 0.304 & 0.275 & 2.35 & 2.47 \\
Match. & transp. & $1.9{\times}10^{9}$ & 14.5 & $1.1{\times}10^{9}$ & 8.43 & \textbf{1.8} & \textbf{1.7} & 0.99 & \textbf{1.0} & 33.3 & 21.5 & 74.5 & 62 \\
Langf. & rev. & $3.0{\times}10^{8}$ & 4.56 & $2.0{\times}10^{8}$ & 3.26 & \textbf{1.6} & \textbf{1.4} & \textbf{1.4} & \textbf{1.1} & 19.4 & 15.7 & 21.3 & 17.3 \\
Mutil. & refl. & $3.3{\times}10^{2}$ & 0.0064 & $3.3{\times}10^{2}$ & 0.0076 & 1.0 & 0.84 & 0.99 & 0.94 & 0.0012 & 0.00162 & 16.3 & 16.4 \\
PHP & vp & $2.4{\times}10^{1}$ & 0.0041 & $3.9{\times}10^{2}$ & 0.0037 & 0.062 & \textbf{1.1} & 0.12 & \textbf{1.1} & 0.000352 & 0.00383 & 0.794 & 0.617 \\
\bottomrule
\end{tabular}}
\caption{Effect of a verified symmetry-breaking constraint (SBC) per family, and UNSAT certificates checking times. SBCs: \emph{vp} value precedence, $x_0{=}0$ first-variable fixing (the two-color case of value precedence), \emph{transp.}\ transposition, \emph{rev.}\ reversal, \emph{refl.}\ reflection. Sizes: Clique $K_3$--$K_{15}$; Myciel.\ $M_2$--$M_4$; Schur $n{=}5$--$45$ (colors $2$--$4$); Odd cyc.\ $C_5$--$C_{51}$; Ramsey $K_6$--$K_{10}$ ($R(3,3)$); vdW $n{=}9$--$28$ (colors $2,3$); Match.\ $K_5$--$K_{19}$; Langf.\ $n{=}2$--$10$; Mutil.\ $4{\times}4$--$12{\times}12$; PHP $n{=}2$--$12$. Columns report, without / with the SBC at the largest instance solved (``lg.''), RoundingSat's \emph{deterministic} operation count and \emph{wall} time (s); the speedups, at that instance and as a geometric mean over the range (``geo.''), with a speedup ${>}1$ in \textbf{bold}; ``Check (native)'', PBLean's compiled checker time (s) on each problem's largest certificate; and ``Check (in-Lean)'', the entire in-Lean checking pipeline against the CSP, also on largest instances. \emph{t/o} marks a 600\,s timeout. Schur's largest instance ($c{=}4$, $n{=}45$) is unsolved without the SBC, so it admits no speedup and is excluded from the geo. means.}
\label{tab:sbc}
\end{table*}

\subsection{Scaling of Pseudo-Boolean versus SAT Proofs}

Before committing to a certification backend, we compared a pseudo-Boolean and a SAT pipeline on a few families. For each problem we generate UNSAT proofs at several instance sizes, solving PB encodings with RoundingSat~\cite{elffersDivideConquer27th2018} (commit \texttt{d4edbf7}, checked by VeriPB 3.0.2) and CNF encodings with CaDiCaL~1.7.3~\cite{biereCaDiCaL202024} (DRAT proofs checked by \texttt{drat-trim}~\cite{wetzlerDRATtrimEfficientChecking2014}), measuring proof length in proof steps.

\Cref{fig:scaling} shows the outcome. On the pigeonhole principle and the mutilated chessboard the pseudo-Boolean proofs stay near-constant while the DRAT proofs grow exponentially, timing out at moderate sizes. This is expected: cutting planes refute the pigeonhole principle compactly~\cite{cookComplexityCuttingplaneProofs1987,hakenIntractabilityResolution1985}, and our cardinality-based ($\sum_i x_i = 1$) mutilated-chessboard encoding admits similar counting arguments~\cite{bryantClausalProofsPseudoBoolean2022,alekhnovichMutilatedChessboardProblem2004}. On odd-cycle non-2-colorability, which is not a counting problem, both proofs scale linearly, suggesting that the gap comes from resolution rather than our encoding.

These observations motivated our choice, rather than dictated it: the pseudo-Boolean pipeline suited the problem examples we had in mind and is the one we develop in full. They are not a verdict against the SAT route, which is a perfectly feasible alternative that we leave as a natural extension for future work.

\subsection{Verified Symmetry Breaking and Pipeline Cost}

As discussed in \Cref{sec:examplesandeffort}, adding a symmetry-breaking constraint to a CSP has a one-time cost in terms of proof effort. Here, we analyze when that cost is repaid in solving speed. Moreover, we measure the cost of our end-to-end verification pipeline in the same experiment. We consider ten different unsatisfiable problem families as our benchmark: we split the graph coloring problem into three parameterized graph families, the clique $K_n$ with $n-1$ colors, the Mycielski graph $M_j$ (obtained from $M_0 = K_2$ by $j$ Mycielskian iterations, so of chromatic number $j+2$) with $j+1$ colors, and the odd cycle $C_{2n+1}$ with 2 colors; upper bounds for Schur, Ramsey, and van der Waerden numbers; perfect matchings for $K_{2m+1}$; Langford pairings for $n \equiv 1,2 \mod 4$; mutilated chessboard for even sizes; and the pigeonhole principle.
For each we solve every instance with and without an SBC, reporting RoundingSat's deterministic operation count
alongside wall time. We check each generated certificate with PBLean v0.3.1 (Lean 4.30.0, Mathlib v4.30.0), timing both the compiled checker alone and the full in-Lean pipeline, which adds elaboration and compilation of the reflected model.
\Cref{tab:sbc} reports the speedups, at the largest instance and as a geometric mean over the range, which we weigh against the one-time proof effort per family of \Cref{tab:proofeffort-loc}. 

Symmetry breaking pays off most where it is theoretically most powerful. Value precedence removes the whole group of permutations of the $k$ interchangeable values, reducing the search space by up to a factor of $k!$, and the measured speedup grows accordingly whenever the base instance requires search. In the clique and Mycielski families the number of colors grows with the instance, and so does the payoff, up to $2\times10^{7}$ on $K_{15}$ and $303\times$ on $M_4$; it is also what makes the four-color Schur instance at $n{=}45$ solvable at all. Where the symmetry group has a fixed size instead, the speedup is bounded and does not grow with the instance, as in the two-color Ramsey ($1.8$--$1.9\times$) and odd cycle ($8.8$--$8.6\times$) families or in Langford's reversal ($1.4$--$1.6\times$).

\looseness=-2
The gain disappears in two cases. First, when the solver needs no search: RoundingSat settles the pigeonhole principle in 24 deterministic operations, so value precedence only adds constraints and slows it, and the mutilated chessboard, dispatched by the same cutting-planes reasoning as above, gives an identical operation count with and without its reflection SBC. Second, when the constraint breaks only a small part of the symmetries: for instance, the matching transposition removes one symmetry out of the factorially many of $K_{2m+1}$.

Wall-clock speedups run consistently closer to one, because wall time cannot fall below a fixed floor of about \SI{3}{\milli\second}. A search space reduction shows in time when the solving time stays above it, and it hides regressions such as the pigeonhole slowdown entirely as well. We therefore read the deterministic operation count as the primary measure of a constraint's effect on search.

Finally, PBLean's compiled checker runs in the same order of magnitude as the solving it certifies (from $0.98\times$ on the clique to $4.3\times$ on Langford), and symmetry breaking shrinks it by the same mechanism, since less search yields a shorter proof: the clique certificate falls from \SI{391}{\mega\byte} to \SI{10}{\kilo\byte} and its checking from \SI{183}{\second} to \SI{5}{\milli\second}. On top of checking, the full pipeline must compile the reflected model, a cost that grows with the size of the encoding rather than with the proof: in comparison, it is minor when the certificate is large (matching: \SI{33}{\second} within \SI{74}{\second}) and dominant when the instance is solved by propagation but its encoding is large, as in mutilated chessboard, whose \SI{1.6}{\milli\second} of checking sit inside \SI{16.4}{\second}. When using SBCs, certifying the largest instance stays under two minutes throughout.

\section{Conclusion}
\label{sec:concl}

\looseness=-2
We presented LeanCSP, a framework for certifying both constraint reformulation and constraint solving inside the Lean theorem prover. It allows one to prove equivalence, equisatisfiability, and symmetry-breaking soundness once for entire parameterized families of CSPs, and to certify the (un)satisfiability of individual instances through translation to external solvers, yielding an end-to-end pipeline that verifies these results inside Lean. 
As the cautionary example of \Cref{sec:cautionary} shows, our framework rejects reformulations whose correctness is not proved.
We hope that researchers will build on LeanCSP to formulate and settle open verification and combinatorial problems with end-to-end formal guarantees.

\av{
  \section*{Acknowledgements}

  This research was funded in whole or in part by the Austrian Science Fund (FWF) 10.55776/COE12.
}

\bibliography{references}

\end{document}